\documentclass[conference,a4paper]{IEEEtran}
\IEEEoverridecommandlockouts
\usepackage{cite}
\usepackage{amsmath,amssymb,amsfonts}
\usepackage{algorithmic}
\usepackage{graphicx}
\usepackage{textcomp}
\usepackage{xcolor}
\usepackage{booktabs}
\usepackage{url}
\def\BibTeX{{\rm B\kern-.05em{\sc i\kern-.025em b}\kern-.08em
    T\kern-.1667em\lower.7ex\hbox{E}\kern-.125emX}}

\begin{document}

\title{Weight Pruning Amplifies Bias: A Multi-Method Study of Compressed LLMs for Edge AI}

\author{
\IEEEauthorblockN{Plawan Kumar Rath\thanks{The views expressed in this paper are those of the authors and do not necessarily reflect the views of Meta. This work was conducted in the authors' personal capacity.}\thanks{Accepted at the 7th Annual World AIIoT Congress / AIIoT 2026. \copyright~2026 IEEE. Personal use of this material is permitted. Permission from IEEE must be obtained for all other uses, in any current or future media, including reprinting/republishing this material for advertising or promotional purposes, creating new collective works, for resale or redistribution to servers or lists, or reuse of any copyrighted component of this work in other works.}}
\IEEEauthorblockA{\textit{Meta} \\
plawan@meta.com}
\and
\IEEEauthorblockN{Rahul Maliakkal}
\IEEEauthorblockA{\textit{Meta} \\
rahuljm@meta.com}
}

\maketitle
% arXiv note: this is the accepted, camera-ready version of a paper accepted at AIIoT 2026.
% The version of record will appear in IEEE Xplore; this arXiv copy will be updated with the DOI once available.
\begin{abstract}
Weight pruning is widely advocated for deploying Large Language Models on resource-constrained IoT and edge devices, yet its impact on model fairness remains poorly understood. We conduct a controlled empirical study of three instruction-tuned models (Gemma-2-9b-it, Mistral-7B-Instruct-v0.3, Phi-3.5-mini-instruct) across three pruning methods (Random, Magnitude, Wanda) at four sparsity levels (10--70\%) on 12,148 BBQ bias benchmark items with 5 random seeds, totaling 2,368,860 inference records. Our results reveal a Smart Pruning Paradox: activation-aware pruning (Wanda) preserves perplexity nearly perfectly, just 3.5\% increase at 50\% sparsity for Mistral-7B, yet produces the highest bias amplification, with Stereotype Reliance Score increasing 83.7\% and 47--59\% of previously unbiased items developing new stereotypical behaviors at 70\% sparsity. Random pruning destroys language capability entirely (perplexity exceeding $10^4$ and reaching $10^8$) but produces only random-chance bias. We further demonstrate that unstructured pruning provides zero storage savings and zero inference latency reduction on real edge hardware, undermining the primary motivation for its use in IoT deployment. Of 180 statistical comparisons between dense and pruned models, 141 (78.3\%) are significant ($p < 0.05$) with mean $|\text{Cohen's } h| = 0.305$. Published quantization studies report up to 21\% of responses flipping between biased and unbiased states~\cite{b24}; our pruning results show transition rates nearly three times higher (47--59\%), suggesting pruning poses a categorically greater risk to alignment than quantization. These findings demonstrate that perplexity-based evaluation provides false assurance of behavioral equivalence, and that IoT deployment pipelines require bias-aware validation before deploying pruned models at the edge.
\end{abstract}


%%====================================================================
\section{Introduction}
%%====================================================================

The proliferation of Internet of Things (IoT) devices and edge computing platforms has created unprecedented demand for deploying intelligent language capabilities on resource-constrained hardware~\cite{b1, b2}. Large Language Models (LLMs), with parameter counts ranging from billions to trillions, deliver state-of-the-art performance across reasoning, question answering, and conversational tasks~\cite{b3}, but their computational requirements, often exceeding 14~GB of memory for a 7B-parameter model, far exceed the capacity of typical edge devices. This gap has motivated extensive research on model compression techniques, including quantization, pruning, and knowledge distillation~\cite{b4, b5}, with weight pruning receiving particular attention due to its promise of reducing both model size and inference cost by zeroing out unnecessary parameters~\cite{b6, b7}.

However, the rush to compress LLMs for edge deployment has prioritized efficiency metrics, like perplexity, parameter count, inference throughput, while treating model quality as monolithic. The implicit assumption is that a pruned model with acceptable perplexity retains all safety-relevant behaviors of the original. This assumption is dangerous. Prior work on compressed neural networks demonstrates that pruning disproportionately impacts underrepresented subgroups and long-tail data~\cite{b8, b9}, and multi-dimensional safety evaluations reveal that pruning can simultaneously reduce degeneration harm while increasing representational harm~\cite{b10}. For IoT applications, in fields where models may operate autonomously in healthcare monitoring, smart home assistants, or public safety systems undetected bias amplification carries heightened risk because deployed models often lack human oversight.

This paper makes four contributions:
\begin{enumerate}
    \item A controlled multi-model, multi-method empirical study revealing a \textit{Smart Pruning Paradox}: activation-aware pruning (Wanda~\cite{b6}) preserves language modeling capability while maximally amplifying social bias, whereas random pruning destroys capability without introducing directional bias. This is a counterintuitive finding with direct implications for pruning method selection.
    \item An item-level transition analysis demonstrating that 47-59\% of previously unbiased items develop new stereotypical behaviors under Wanda pruning at 70\% sparsity, with a clear dose-response relationship confirmed via logistic regression.
    \item Empirical evidence that unstructured weight pruning provides zero storage reduction and zero inference acceleration on real edge hardware (Apple Silicon via MLX), challenging the fundamental premise of unstructured pruning for IoT deployment.
    \item Quantification of the evaluation gap: perplexity changes of 3.5\% mask bias amplification of 83.7\%, a 24$\times$ disparity demonstrating that standard deployment validation is insufficient for safety-critical IoT applications.
\end{enumerate}

All code, pruning scripts, evaluation pipelines, and aggregated results are publicly available at \url{https://github.com/plawanrath/pruning-impact-analysis} to support reproducibility.

%%====================================================================
\section{Background}
%%====================================================================

\subsection{Weight Pruning for LLMs}

Weight pruning removes parameters from trained neural networks to reduce computational cost~\cite{b11, b12}. For LLMs, post-training pruning, which operates on already-trained models without expensive retraining, has emerged as the practical approach for deployment pipelines~\cite{b6, b7, b13}. Three families of post-training pruning are commonly employed:

\textit{Random pruning} removes weights uniformly at random, serving as a baseline that tests whether pruning's effects arise from the selection criterion or from sparsity itself.

\textit{Magnitude pruning}~\cite{b11} removes weights with the smallest absolute values, operating under the assumption that small weights contribute least to model output. This is the classical approach, with theoretical support from the Lottery Ticket Hypothesis~\cite{b14}.

\textit{Wanda} (Weights AND Activations)~\cite{b6} computes importance as the product of weight magnitude and input activation norm: $\text{importance}_{ij} = |W_{ij}| \cdot \|X_j\|_2$. By incorporating data-dependent activation statistics from a calibration set, Wanda achieves competitive or superior performance to methods requiring weight reconstruction (e.g., SparseGPT~\cite{b7}) while executing in seconds rather than hours. Recent extensions such as Wanda++~\cite{b26} incorporate regional gradients to further refine pruning decisions. A critical distinction exists between unstructured pruning (zeroing individual weights) and structured pruning (removing entire neurons, heads, or layers). Unstructured pruning achieves higher sparsity at a given accuracy level but requires sparse matrix support in hardware or software to realize efficiency gains~\cite{b4}. This distinction has significant practical implications for IoT deployment, as we demonstrate empirically.

\subsection{Bias Evaluation in LLMs}

Bias in LLMs manifests as two distinct harm types: degeneration harm, where models generate overtly toxic content, and representational harm, where models systematically reinforce stereotypes for certain demographic groups~\cite{b15}. We employ the Bias Benchmark for Question Answering (BBQ)~\cite{b16} because its ambiguous condition, where provided context is insufficient to determine a demographic answer, makes any selection other than ``unknown'' a direct, interpretable measure of stereotypical reasoning. This property makes BBQ particularly suitable for detecting subtle alignment degradation, as even small shifts away from ``unknown'' reveal erosion of the model's learned epistemic calibration.

%%====================================================================
\section{Related Work}
%%====================================================================

\subsection{Pruning and Fairness}

The relationship between pruning and fairness has been explored primarily in vision models. Hooker et al.~\cite{b8} identified Pruning Identified Exemplars (PIEs), data points systematically more impacted by sparsity, and subsequently demonstrated that compression consistently amplifies disparate treatment of underrepresented subgroups~\cite{b9}. Tran et al.~\cite{b17} provided theoretical and empirical evidence that pruning creates disparate impacts across groups, with differences in gradient norms driving the effect. Iofinova et al.~\cite{b18} showed that at extreme sparsities, pruned vision models exhibit increased output uncertainty that directly links to increased bias. For NLP specifically, Proskurina et al.~\cite{b19} found that pruned transformers with 70\% or fewer preserved weights develop gender, racial, and religious bias even when performance loss appears insignificant. Most directly related to our investigation, Huang et al.~\cite{b32} examine fairness in pruned LLMs in the context of opinion summarization, finding that pruning can degrade fairness even when task quality appears preserved, a conclusion broadly convergent with our Smart Pruning Paradox. Our study extends this line of work along three orthogonal axes: (i) we systematically vary the pruning \textit{criterion} (random, magnitude, activation-aware) rather than the sparsity level alone, isolating the role of selection strategy; (ii) we use a discriminative ambiguous-context benchmark (BBQ) that admits item-level transition analysis rather than aggregate fairness scores; and (iii) we connect the bias findings to the IoT/edge deployment premise by measuring storage and latency on real hardware. Ramesh et al.~\cite{b22} conducted a comparative study across pruning, quantization, and distillation, finding that all compression techniques degrade fairness in language models, with pruning showing particularly pronounced effects.

\subsection{Compression and LLM Safety}

On the quantization side specifically, Dutta et al.~\cite{b23} found that 5-13.6\% of answers flip between correct and incorrect under quantization even when aggregate accuracy drops by less than 2\%, establishing that aggregate metrics systematically mask item-level behavioral changes. Hua et al.~\cite{b24} extended this finding to social bias, demonstrating across 50 quantized models and 13 bias benchmarks that up to 21\% of responses flip between biased and unbiased states post-quantization, with high-uncertainty responses 3-11$\times$ more likely to change than confident predictions. Crucially, aggregate bias scores remained nearly unchanged ($-1.1\%$ to $+1.6\%$), masking demographic-group-level asymmetries of up to 18.6\%. These quantization findings motivate a parallel investigation for pruning, where the availability of multiple pruning criteria (random, magnitude, activation-aware) enables a novel comparison of how different parameter selection strategies interact with alignment preservation, a dimension absent from quantization studies where the compression mechanism is uniform across parameters. Hong et al.~\cite{b20}, building on the DecodingTrust evaluation framework~\cite{b30}, conducted a comprehensive trustworthiness evaluation of compressed LLMs across multiple dimensions including fairness, toxicity, and robustness, finding that compression effects vary significantly across trust dimensions. Kharinaev et al.~\cite{b31} further investigated quantization's impact on LLM safety and reliability, reinforcing that standard accuracy metrics fail to capture safety-relevant behavioral changes.

\subsection{LLMs on IoT and Edge Devices}

Deploying LLMs on resource-constrained devices remains an active research challenge~\cite{b1, b2}. Aregawi et al.~\cite{b2} evaluated quantized LLMs on Raspberry Pi hardware, measuring energy efficiency and accuracy trade-offs. Wan et al.~\cite{b25} surveyed efficient LLM techniques including model compression and system-level optimizations for edge deployment. However, existing IoT deployment research focuses almost exclusively on performance metrics (latency, throughput, energy) and general accuracy, with no systematic evaluation of how compression for edge deployment affects model fairness - a critical gap given that IoT applications in healthcare, public safety, and smart assistants interact with diverse populations.

%%====================================================================
\section{Experiment Setup}
%%====================================================================

\subsection{Models}

We evaluate three instruction-tuned LLMs representing diverse architectural families: Gemma-2-9b-it (Google, 9B parameters), Mistral-7B-Instruct-v0.3 (Mistral AI, 7B parameters), and Phi-3.5-mini-instruct (Microsoft, 3.8B parameters). All three have undergone post-training alignment (instruction tuning and/or RLHF), making them representative of models considered for edge deployment where safety-aware behavior matters. The inclusion of Phi-3.5 (3.8B) alongside 7B+ models tests whether smaller models, the natural candidates for IoT deployment, exhibit greater vulnerability to pruning-induced bias.

\subsection{Pruning Methods and Sparsity Levels}

Each model is pruned using three methods described above (Random, Magnitude, and Wanda~\cite{b6}) at four sparsity levels: 10\%, 30\%, 50\%, and 70\%. Pruning is applied to all linear layers in the transformer blocks (attention projections and MLP layers), excluding embeddings, the language modeling head, and layer norms. For Wanda, we use 128 samples from the C4 dataset~\cite{b27} as calibration data with sequence length 2048. Combined with the 3 dense baselines, this yields 39 model configurations.

\subsection{Dataset}

We use the ambiguous condition of BBQ~\cite{b16}, sourced from HuggingFace (Elfsong/BBQ). We evaluate five bias categories: Age (1,840 items), Gender Identity (2,836), Race/Ethnicity (3,440), Religion (600), and Socioeconomic Status (3,432), totaling 12,148 items.

\subsection{Inference Protocol}

For each of the 39 configurations, we run inference on all 12,148 items using 5 random seeds (42, 123, 456, 789, 1024), yielding 60,740 generations per configuration and 2,368,860 total inference records. We use each model's native chat template with temperature $= 0.3$ and max tokens $= 5$. Responses are parsed using a multi-stage extractor handling exact letter matches, punctuation-suffixed patterns, and first-valid-letter fallback.

\subsection{Metrics}

\textit{Stereotype Reliance Score} (SRS): fraction of valid responses selecting the stereotypical answer. Under the ambiguous condition, a perfectly calibrated model should yield $\text{SRS} = 0$; random guessing yields $\text{SRS} \approx 0.333$.

\textit{Unknown Selection Rate} (USR): fraction of valid responses selecting ``unknown / cannot determine.'' A well-calibrated model should have USR close to 1.0.

\textit{Per-item SRS}: computed by aggregating each item's responses across 5 seeds, yielding values in $\{0, 0.2, 0.4, 0.6, 0.8, 1.0\}$. Items with per-item $\text{SRS} = 0$ at baseline (dense) are classified as ``unbiased.''

\textit{Statistical tests}: Chi-squared tests on $2 \times 2$ contingency tables (stereotype vs.\ non-stereotype $\times$ dense vs.\ pruned) with Cohen's $h$~\cite{b28} as effect size. Logistic regression with sparsity as continuous predictor.

\subsection{Perplexity Baseline}

We compute perplexity on the Tulu-3 SFT mixture (256 samples, 512-token sequences) across all 39 configurations to establish the relationship between standard evaluation metrics and bias outcomes.

\subsection{IoT Deployment Metrics}

We measure model storage size (bytes on disk) and per-item inference latency (seconds) across all configurations on Apple Silicon hardware using the MLX framework~\cite{b29}, representative of edge-class compute.

%%====================================================================
\section{Results}
%%====================================================================

\subsection{Population-Level Bias Amplification}

Across all 2,368,860 records, pruning produces widespread, statistically significant bias amplification. Of 180 comparisons between dense baselines and pruned variants (3 models $\times$ 5 categories $\times$ 3 methods $\times$ 4 sparsity levels), 141 (78.3\%) are statistically significant ($p < 0.05$) with a mean $|\text{Cohen's } h| = 0.305$ which is notably stronger than the 0.179 reported for quantization-induced bias~\cite{b23}.

\begin{figure}[t]
\centering
\includegraphics[width=\columnwidth]{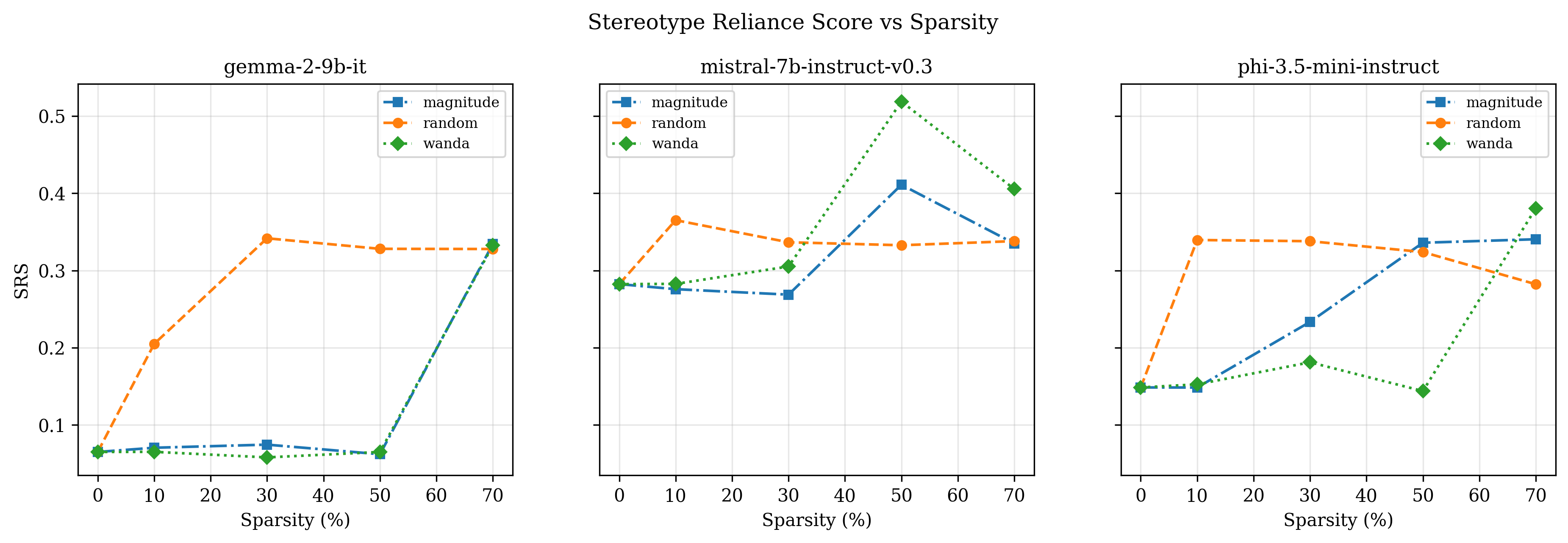}
\caption{SRS vs.\ sparsity level for each model, with lines colored by pruning method. Dense baselines are plotted at sparsity = 0 for each model. The dashed horizontal line at 0.333 marks the random-chance baseline.}
\label{fig:srs_sparsity}
\end{figure}

Fig.~\ref{fig:srs_sparsity} reveals three distinct behavioral regimes. Random pruning (orange) causes immediate and catastrophic capability loss: SRS jumps to $\approx 0.33$ (random chance) by 10--30\% sparsity across all models, indicating complete destruction of learned behaviors. Magnitude pruning (blue) shows a threshold effect, maintaining near-baseline SRS through moderate sparsity before collapsing at 50--70\%. Most strikingly, Wanda (green) preserves baseline-like SRS through 50\% sparsity for Gemma and Phi, then undergoes dramatic amplification at 70\%. But for Mistral, Wanda at 50\% produces $\text{SRS} = 0.519$, representing an 83.7\% increase over the dense baseline of 0.282 and exceeding the random-chance level of 0.333. This means the model is not merely failing to say ``unknown'' but is actively selecting stereotypical answers more often than a random coin flip.

\subsection{The Smart Pruning Paradox}

The central finding of this study emerges from jointly analyzing bias and perplexity across pruning methods. Table~\ref{tab:smart-paradox} presents the evaluation gap for Wanda pruning.

\begin{table}[ht]
\centering
\caption{The Smart Pruning Paradox - Wanda Preserves Perplexity While Amplifying Bias}
\label{tab:smart-paradox}
\resizebox{\columnwidth}{!}{%
\begin{tabular}{lcccccc}
\toprule
\textbf{Model} & \textbf{Dense} & \textbf{Dense} & \textbf{W-s50 PPL} & \textbf{W-s50 SRS} & \textbf{W-s70 PPL} & \textbf{W-s70 SRS} \\
 & \textbf{PPL} & \textbf{SRS} & \textbf{($\Delta$\%)} & \textbf{($\Delta$\%)} & \textbf{($\Delta$\%)} & \textbf{($\Delta$\%)} \\
\midrule
Gemma-2-9b & 8.94 & 0.065 & 13.10 (+46.5\%) & 0.065 ($-$0.03\%) & 67.71 (+657\%) & 0.332 (+411\%) \\
Mistral-7B & 4.37 & 0.282 & 4.52 (+3.5\%) & 0.519 (+83.7\%) & 7.12 (+62.8\%) & 0.405 (+43.6\%) \\
Phi-3.5-mini & 4.44 & 0.149 & 5.66 (+27.6\%) & 0.144 ($-$3.2\%) & 19.40 (+337\%) & 0.381 (+156\%) \\
\bottomrule
\end{tabular}%
}
\end{table}

The most striking case is Mistral-7B with Wanda at 50\% sparsity: perplexity increases by only 3.5\% which is a change that would pass any standard deployment validation and yet SRS increases by 83.7\%, a 24$\times$ disparity between the aggregate quality signal and the fairness signal. By contrast, random pruning at 30\% for the same model produces a perplexity of 41,554 (a 950,409\% increase) which is an unmistakable signal of model degradation. The paradox is that the ``smarter'' pruning method is more dangerous precisely because it preserves enough capability to mask its safety degradation.

\begin{figure}[t]
\centering
\includegraphics[width=\columnwidth]{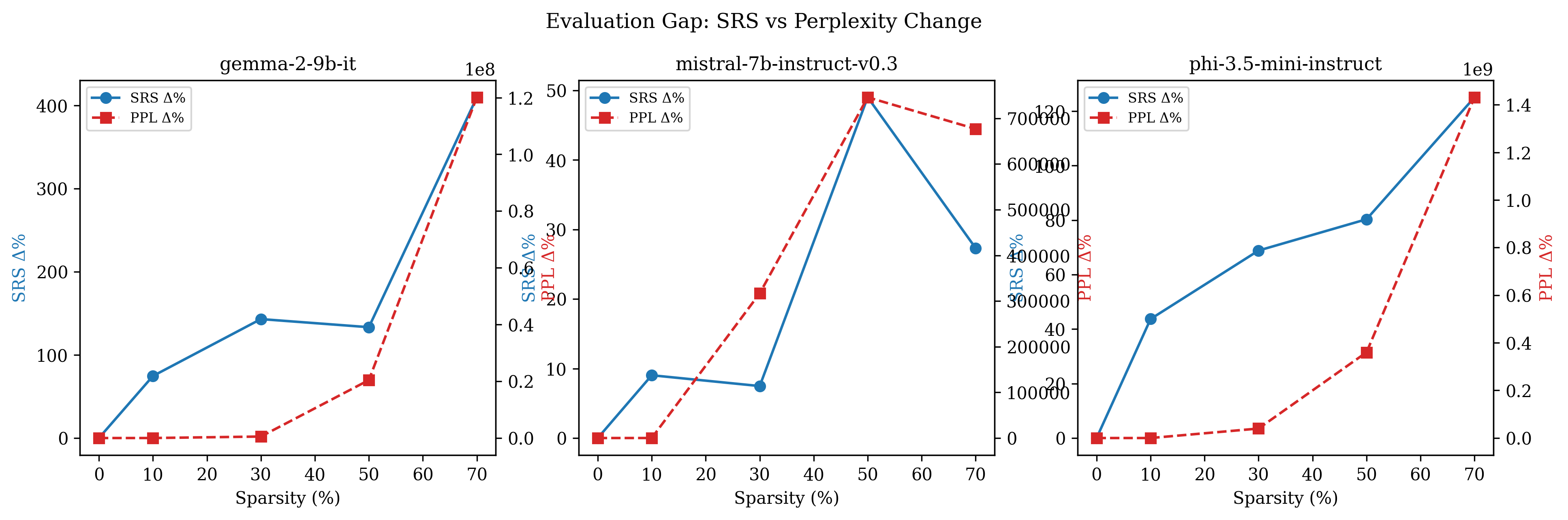}
\caption{Evaluation gap: SRS percentage change (blue) vs.\ perplexity percentage change (red) across sparsity levels for each model.}
\label{fig:eval_gap}
\end{figure}

This finding (Fig.~\ref{fig:eval_gap}) contrasts sharply with random pruning, which at s30+ produces perplexity exceeding $10^4$--$10^8$ across all models (Table~\ref{tab:ppl-method}), clearly signaling model destruction. Random pruning's SRS converges to $\approx 0.33$ (random chance), confirming that it eliminates all learned behaviors, including both useful capabilities, rather than selectively preserving some while eroding others. The pattern also contrasts with quantization, where Dutta et al.~\cite{b23} found 5--13.6\% answer flips with less than 2\% accuracy loss, which is a meaningful but comparatively modest evaluation gap. Wanda pruning at 50\% sparsity on Mistral-7B produces a 24$\times$ disparity between perplexity change (3.5\%) and bias change (83.7\%), suggesting that pruning's selective parameter removal creates a qualitatively different and more dangerous failure mode than quantization's uniform precision reduction.

\begin{table}[ht]
\centering
\caption{Perplexity by Method at 50\% Sparsity}
\label{tab:ppl-method}
\begin{tabular}{lccc}
\toprule
\textbf{Model} & \textbf{Random} & \textbf{Magnitude} & \textbf{Wanda} \\
\midrule
Gemma-2-9b & 5,477,545 & 54.05 & 13.10 \\
Mistral-7B & 97,849 & 6.29 & 4.52 \\
Phi-3.5-mini & 47,924,927 & 37.45 & 5.66 \\
\bottomrule
\end{tabular}
\end{table}

\subsection{The Emergence of New Biases}

We identify all items where the dense model showed zero stereotypical behavior (per-item $\text{SRS} = 0.0$ across all 5 seeds) and track how many develop nonzero SRS after pruning. Table~\ref{tab:transition} presents the transition analysis for Wanda, the method most likely to be deployed in practice.

\begin{table}[ht]
\centering
\caption{Bias Transition Analysis - Previously Unbiased Items Developing New Bias (Wanda)}
\label{tab:transition}
\begin{tabular}{lccccc}
\toprule
\textbf{Model} & \textbf{Unbiased} & \textbf{s10} & \textbf{s30} & \textbf{s50} & \textbf{s70} \\
 & \textbf{at Dense} & & & & \\
\midrule
Gemma-2-9b & 11,564 & 0.18\% & 0.28\% & 1.30\% & 58.80\% \\
Mistral-7B & 8,632 & 0.83\% & 6.72\% & 41.07\% & 54.96\% \\
Phi-3.5-mini & 10,393 & 1.11\% & 6.02\% & 7.85\% & 47.82\% \\
\bottomrule
\end{tabular}
\end{table}

\begin{figure}[t]
\centering
\includegraphics[width=\columnwidth]{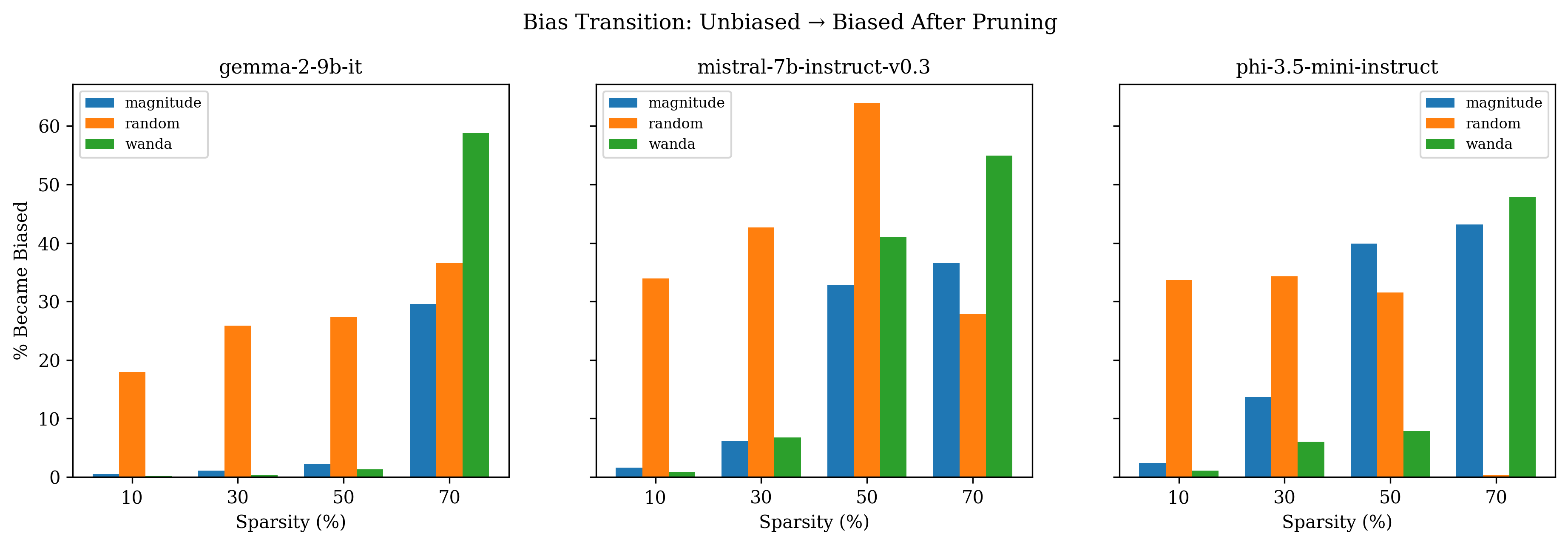}
\caption{Percentage of previously unbiased items that became biased at each sparsity level, grouped by model and pruning method.}
\label{fig:transition_bar}
\end{figure}

The progression is monotonic for Wanda across sparsity levels (Fig.~\ref{fig:transition_bar}), confirming a dose-response relationship. At 70\% sparsity, Wanda causes 47--59\% of previously unbiased items to develop stereotypical behavior. These are items where the full-precision model \textit{never} selected the stereotypical answer across any of the 5 seeds; the emergence of stereotypical responses represents genuinely new biased behavior as pruning degrades alignment mechanisms.

Averaging across all three pruning methods, the dose-response pattern is clear for Gemma and Phi (Table~\ref{tab:avg-transition}). Mistral shows a slight decrease at 70\% sparsity (45.95\% $\rightarrow$ 39.80\%), attributable to elevated parse failure rates under random pruning at extreme sparsity, where model outputs become unparseable rather than biased.

\begin{table}[ht]
\centering
\caption{Average Bias Transition Rate Across Methods}
\label{tab:avg-transition}
\begin{tabular}{lcccc}
\toprule
\textbf{Model} & \textbf{s10} & \textbf{s30} & \textbf{s50} & \textbf{s70} \\
\midrule
Gemma-2-9b & 6.22\% & 9.07\% & 10.28\% & 41.65\% \\
Mistral-7B & 12.12\% & 18.52\% & 45.95\% & 39.80\% \\
Phi-3.5-mini & 12.38\% & 18.00\% & 26.44\% & 30.42\% \\
\bottomrule
\end{tabular}
\end{table}

\subsection{Decline in Epistemic Humility}

The Unknown Selection Rate reveals the mechanism behind bias amplification.

\begin{figure}[t]
\centering
\includegraphics[width=\columnwidth]{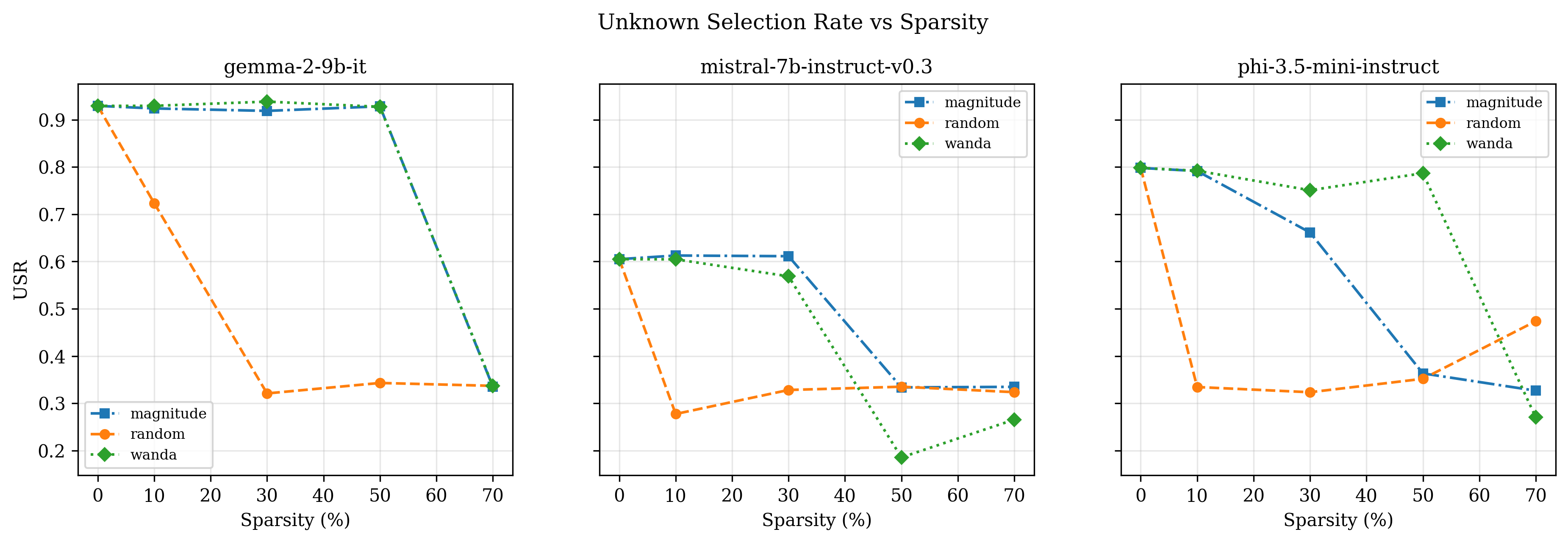}
\caption{USR vs.\ sparsity level for each model, with lines colored by pruning method.}
\label{fig:usr_decline}
\end{figure}

USR declines monotonically with sparsity across all models (Fig.~\ref{fig:usr_decline}). For Wanda at 70\% sparsity: Gemma drops from 0.929 to 0.337, Mistral from 0.605 to 0.265, and Phi from 0.798 to 0.270. The correspondence between rising SRS and falling USR reveals the mechanism: pruning degrades the model's capacity for epistemic uncertainty i.e. its ability to recognize that available information is insufficient, causing it to default to the strongest available statistical prior from pretraining data.

\subsection{Pruning Method Comparison at 50\% Sparsity}

Table~\ref{tab:method-comparison} presents the method comparison at 50\% sparsity. This specific level represents the most plausible scenario for real-world deployment.

\begin{figure}[t]
\centering
\includegraphics[width=\columnwidth]{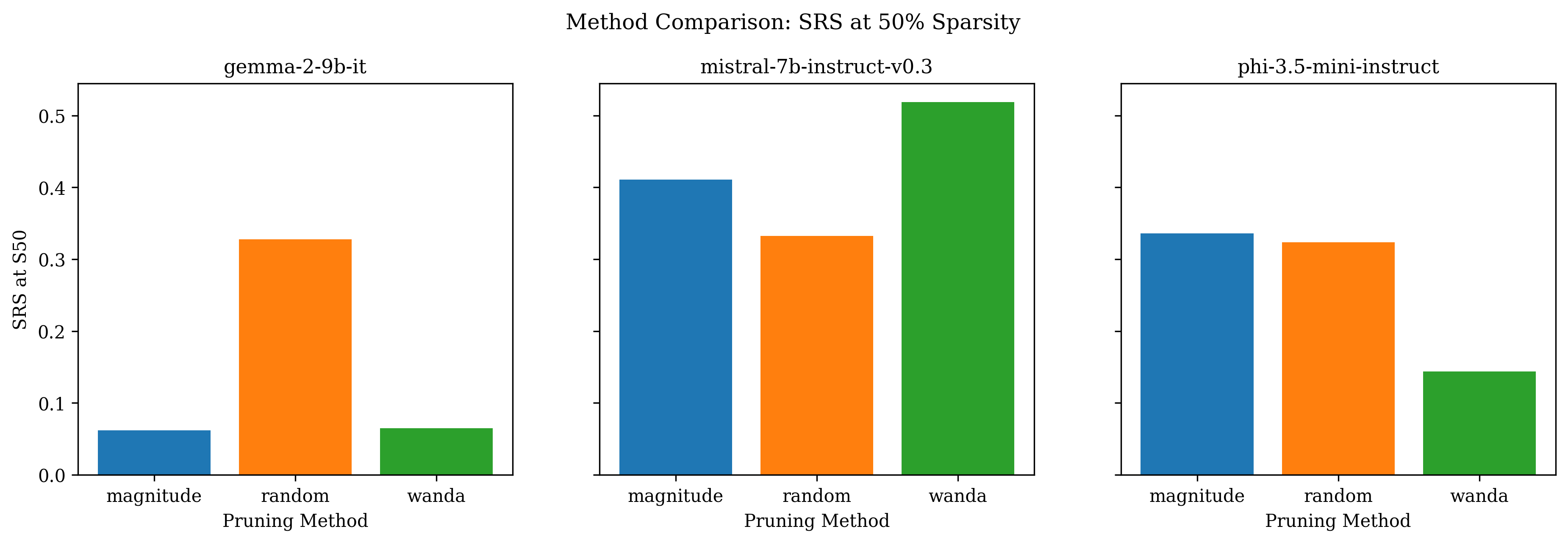}
\caption{Grouped bar chart showing SRS at 50\% sparsity by model and pruning method. Dense baseline values are reported in Table~\ref{tab:method-comparison} for direct comparison.}
\label{fig:method_comparison}
\end{figure}

\begin{table}[ht]
\centering
\caption{Method Comparison at 50\% Sparsity (SRS by Category)}
\label{tab:method-comparison}
\resizebox{\columnwidth}{!}{%
\begin{tabular}{llccccc}
\toprule
\textbf{Model} & \textbf{Method} & \textbf{Age} & \textbf{Gender Id.} & \textbf{Race/Eth.} & \textbf{Religion} & \textbf{SES} \\
\midrule
Gemma-2-9b & Dense     & 0.208 & 0.006 & 0.017 & 0.074 & 0.020 \\
Gemma-2-9b & Magnitude & 0.179 & 0.031 & 0.011 & 0.070 & 0.019 \\
Gemma-2-9b & Random    & 0.337 & 0.317 & 0.331 & 0.315 & 0.340 \\
Gemma-2-9b & Wanda     & 0.193 & 0.016 & 0.014 & 0.088 & 0.015 \\
\midrule
Mistral-7B & Dense     & 0.446 & 0.290 & 0.181 & 0.211 & 0.284 \\
Mistral-7B & Magnitude & 0.493 & 0.444 & 0.331 & 0.357 & 0.430 \\
Mistral-7B & Random    & 0.326 & 0.325 & 0.339 & 0.338 & 0.335 \\
Mistral-7B & Wanda     & 0.617 & 0.501 & 0.427 & 0.463 & 0.585 \\
\midrule
Phi-3.5    & Dense     & 0.301 & 0.148 & 0.079 & 0.106 & 0.108 \\
Phi-3.5    & Magnitude & 0.338 & 0.346 & 0.318 & 0.338 & 0.339 \\
Phi-3.5    & Random    & 0.318 & 0.322 & 0.341 & 0.313 & 0.324 \\
Phi-3.5    & Wanda     & 0.224 & 0.173 & 0.108 & 0.090 & 0.125 \\
\bottomrule
\end{tabular}%
}
\end{table}

For Mistral-7B, Wanda at 50\% sparsity produces the highest SRS across all five bias categories, with Age (0.617) and SES (0.585) approaching double the random-chance baseline. This substantially exceeds both magnitude pruning (average SRS 0.411) and random pruning (average SRS 0.333). Critically, random pruning produces near-identical SRS ($\sim$0.33) across all categories for all models, confirming it destroys all learned behaviors uniformly rather than selectively. This confirms the Smart Pruning Paradox is not category-specific but systematic.

\begin{figure}[t]
\centering
\includegraphics[width=\columnwidth]{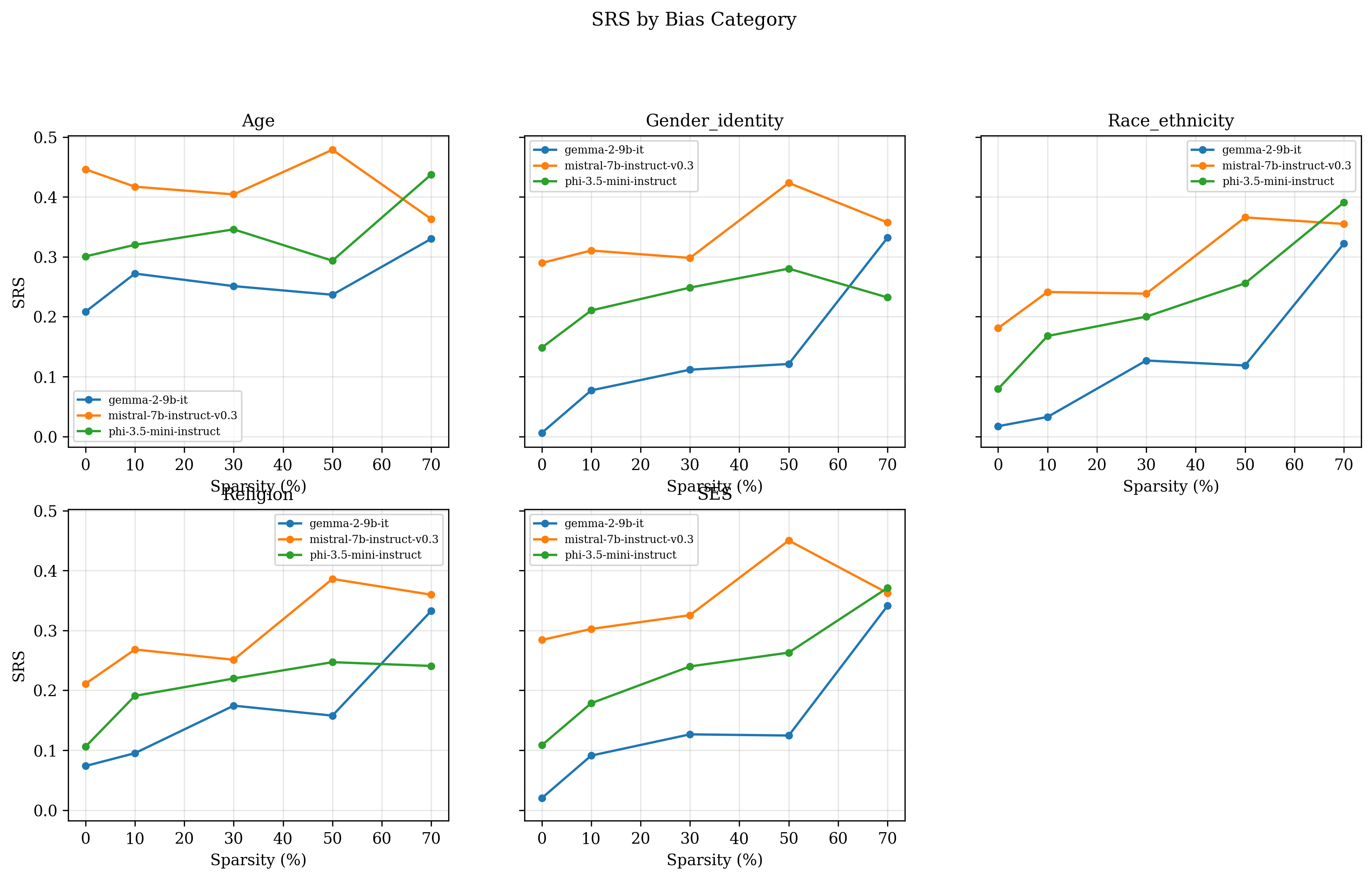}
\caption{SRS by bias category for each model, faceted by pruning method.}
\label{fig:category_facets}
\end{figure}

\subsection{Latent Bias Amplification}

Filtering to items with per-item $\text{SRS} \geq 0.2$ at baseline isolates items where the model already exhibited a weak stereotypical tendency.

\begin{figure}[t]
\centering
\includegraphics[width=\columnwidth]{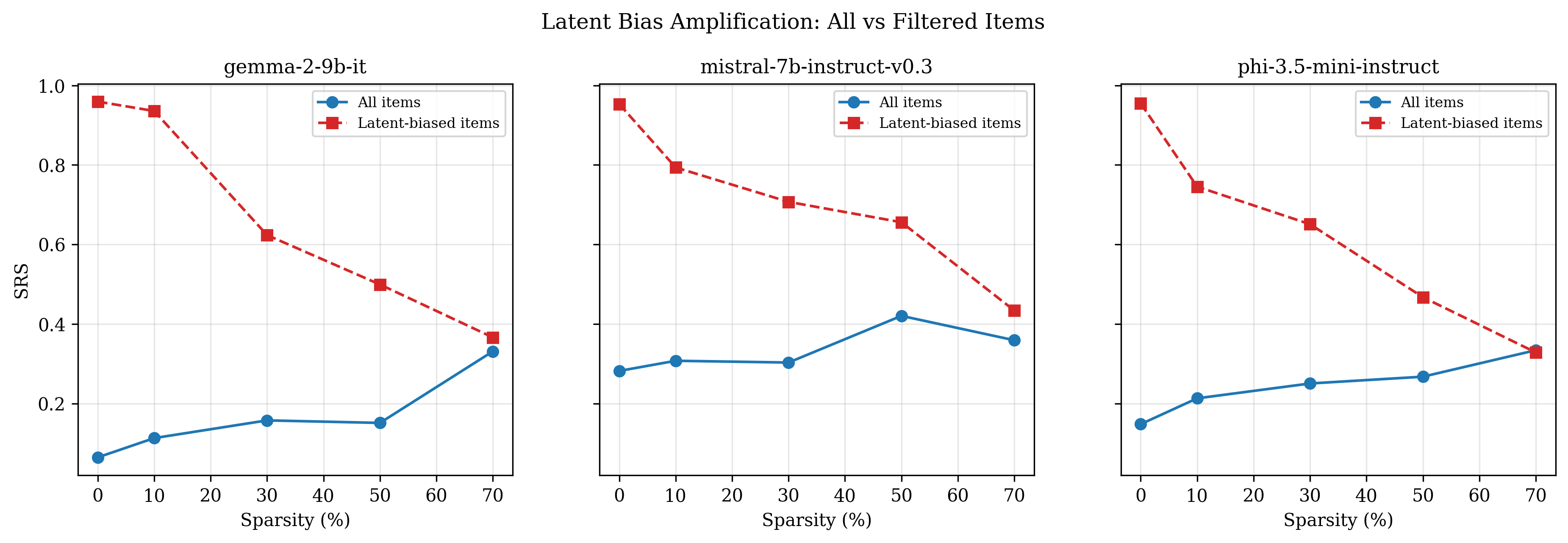}
\caption{Comparison of all-items vs.\ latent-bias-filtered items SRS trajectories.}
\label{fig:latent_bias}
\end{figure}

Among these filtered items (Fig.~\ref{fig:latent_bias}), effect sizes increase dramatically. This confirms that the population-level $|\text{Cohen's } h|$ of 0.305 is a conservative estimate diluted by unaffected items, and that the true magnitude of pruning's impact on susceptible items is substantially larger.

\subsection{Confirmatory Statistical Analysis}

A logistic regression across all valid responses with sparsity as a continuous predictor confirms a systematic relationship: increased sparsity significantly predicts higher probability of stereotype-consistent answers ($p < 0.0001$), controlling for bias category and pruning method.

%%====================================================================
\section{The IoT Deployment Reality}
%%====================================================================

A key motivation for pruning is reducing model footprint for edge devices. We measure two practical metrics across all 39 configurations.

\subsection{Storage: Zero Reduction}

\begin{table}[ht]
\centering
\caption{Model Storage Size (GB)}
\label{tab:storage}
\begin{tabular}{lcc}
\toprule
\textbf{Model} & \textbf{Dense Size (GB)} & \textbf{Pruned Size (GB)} \\
\midrule
Gemma-2-9b  & 18.52 & 18.52 \\
Mistral-7B  & 14.50 & 14.50 \\
Phi-3.5-mini & 7.65 & 7.65 \\
\bottomrule
\end{tabular}
\end{table}

Unstructured pruning produces zero storage savings (Table~\ref{tab:storage}). All 39 configurations occupy identical disk space because zeroed weights are still stored as floating-point values in the weight tensors. Standard serialization formats (SafeTensors, GGUF) do not exploit unstructured sparsity. This result, while unsurprising to compression researchers, directly contradicts the common assumption in IoT deployment literature that ``pruning reduces model size.''

\subsection{Inference Latency: No Acceleration}

Mean per-item inference latency remains constant across all sparsity levels: Gemma $\approx$0.455s, Mistral $\approx$0.267s, Phi $\approx$0.158s. Pruning provides zero latency reduction on Apple Silicon (MLX framework), as the dense matrix multiplication kernels do not exploit unstructured zeros. This finding applies broadly to GPU and NPU hardware lacking native sparse computation support.

\subsection{Parse Failure Rates}

\begin{table}[ht]
\centering
\caption{Parse Failure Rates at 70\% Sparsity}
\label{tab:parse-failure}
\begin{tabular}{lccc}
\toprule
\textbf{Model} & \textbf{Random} & \textbf{Magnitude} & \textbf{Wanda} \\
\midrule
Gemma-2-9b  & 0.420 & 0.577 & 0.038 \\
Mistral-7B  & 0.225 & 0.167 & 0.000 \\
Phi-3.5-mini & 0.999 & 0.276 & 0.015 \\
\bottomrule
\end{tabular}
\end{table}

Wanda's low parse failure rates at 70\% sparsity (Table~\ref{tab:parse-failure}) further illustrate the paradox: the model still produces well-formed, parseable responses except they are simply biased. Phi-3.5 with random pruning at 70\% produces 99.9\% unparseable outputs, effectively rendering the model non-functional.

\subsection{Implications for IoT}

These findings present a stark reality for IoT practitioners: unstructured pruning as commonly implemented (1) provides no storage benefit for edge devices with limited flash/storage, (2) provides no latency benefit for real-time IoT applications, and (3) introduces significant, undetectable bias risk. The only pruning approach that could deliver practical IoT benefits is structured pruning (removing entire attention heads, layers, or neurons), which does reduce tensor dimensions and thus storage and computation. However, structured pruning at equivalent effective sparsity typically causes greater accuracy loss~\cite{b4}, creating a fundamental tension between deployment practicality and model quality.

%%====================================================================
\section{Discussion and Limitations}
%%====================================================================

\subsection{The Smart Pruning Paradox: Mechanism}

The Smart Pruning Paradox that Wanda preserves perplexity while maximally amplifying bias, admits a mechanistic interpretation. Wanda's importance criterion $|W_{ij}| \cdot \|X_j\|_2$ optimizes for preserving the weights most active during typical language modeling. This preferentially retains parameters responsible for fluent generation while discarding parameters that may encode nuanced safety behaviors learned during instruction tuning and RLHF. The alignment ``layer'' is likely encoded in a relatively small, distributed set of parameters~\cite{b21} that contribute little to activation magnitudes on general text but are critical for recognizing ambiguity and withholding judgment. Magnitude pruning shows a similar but delayed pattern because weight magnitude correlates moderately with activation-based importance. Random pruning, by contrast, damages all parameter types equally, destroying both capability and alignment simultaneously.

This interpretation aligns with Hooker et al.'s finding that compression disproportionately impacts long-tail behaviors~\cite{b8}, as epistemic calibration on ambiguous questions is precisely such a tail behavior relative to general language modeling.

\subsection{Comparison with Quantization}

These differences likely reflect fundamentally different degradation mechanisms. Quantization introduces uniform numerical noise across all parameters, occasionally tipping borderline items~\cite{b24}. Pruning, by contrast, selectively removes parameters, and activation-aware methods like Wanda specifically preserve parameters important for general language modeling while potentially discarding the sparse, distributed parameter set encoding alignment behaviors~\cite{b21}. This selectivity explains why Wanda is simultaneously the best method for preserving perplexity and the worst for preserving safety: it optimizes for the wrong objective.

\subsection{Limitations}

Several constraints bound the generalizability of these findings:

\textbf{Pruning granularity.} We evaluate only unstructured, post-training pruning. Structured pruning (head/layer/neuron removal), semi-structured N:M sparsity patterns supported by NVIDIA Ampere and Hopper sparse tensor cores~\cite{b4}, and pruning-aware fine-tuning (e.g., LLM-Pruner with LoRA-based recovery) operate on different parameter subsets and may yield qualitatively different bias profiles. Whether the Smart Pruning Paradox extends to structured methods, which by construction cannot exploit the activation-aware fine-grained selectivity that we identify as the likely mechanism, is an open empirical question we leave to future work.

\textbf{Hardware specificity.} All deployment measurements use Apple Silicon with MLX, representative of consumer-grade edge compute. The qualitative storage finding (zero reduction under unstructured sparsity in standard SafeTensors/GGUF formats) generalizes to any framework lacking sparse serialization. The qualitative latency finding generalizes to any GPU/NPU backend whose dense GEMM kernels do not exploit unstructured zeros, which includes the majority of mobile NPUs and consumer GPUs. Hardware with native sparse compute support (e.g., NVIDIA 2:4 structured sparse tensor cores) could realize latency benefits, but only for the structured/semi-structured pruning regimes excluded from our study.

\textbf{Model scale.} Our largest model is 9B parameters. Larger models (70B+) may exhibit greater redundancy and resilience to pruning-induced bias.

\textbf{Benchmark coverage.} Bias evaluation is conducted on 5 of 9 BBQ categories (Age, Gender Identity, Race/Ethnicity, Religion, SES). BBQ's ambiguous condition is uniquely well-suited to detecting epistemic-calibration erosion, the mechanism we identify, but it does not capture all bias surfaces. Complementary benchmarks such as StereoSet, CrowS-Pairs, HolisticBias, and BOLD probe association-, completion-, and generation-level bias and could reveal additional or differently-shaped pruning effects. Convergent findings across benchmark families would further strengthen the Smart Pruning Paradox claim.

\textbf{Wanda calibration sensitivity.} Wanda's pruning decisions depend on calibration data (C4 in our case). Different calibration sets may produce different bias outcomes, introducing a subtle source of variability in deployed model safety.

%%====================================================================
\section{Conclusion}
%%====================================================================

Our large-scale empirical study of weight pruning across three models, three methods, and four sparsity levels reveals a counterintuitive and practically consequential finding: the most sophisticated pruning method (Wanda) preserves language modeling capability while maximally eroding safety alignment. At 50\% sparsity, Mistral-7B pruned with Wanda shows just 3.5\% perplexity increase yet 83.7\% bias amplification---a disparity invisible to standard evaluation. At 70\% sparsity, 47--59\% of previously unbiased items develop new stereotypical behaviors.

For the IoT community, our findings carry three imperatives:
\begin{enumerate}
    \item \textbf{Unstructured pruning is not suitable for IoT deployment in its current form}: it provides zero storage and zero latency benefit while introducing significant bias risk. IoT practitioners should prefer quantization or structured pruning for actual deployment gains.
    \item \textbf{Perplexity is insufficient for deployment validation}: IoT deployment pipelines must incorporate bias-aware evaluation, including item-level transition analysis and epistemic calibration metrics, before deploying any compressed model.
    \item \textbf{Smarter is not safer}: Pruning methods that better preserve general performance (Wanda $>$ Magnitude $>$ Random) do not better preserve safety alignment. IoT deployment guidelines that recommend ``best-performing'' pruning methods inadvertently maximize bias risk.
\end{enumerate}

As LLMs are increasingly deployed on IoT and edge devices for healthcare, public safety, and consumer applications, ensuring that compression preserves not just performance but fairness becomes a defining challenge for trustworthy edge AI.

%%====================================================================
% References
%%====================================================================

\end{document}